\documentclass[letterpaper, 10 pt, conference]{ieeeconf}  
\IEEEoverridecommandlockouts                             
\usepackage{caption}
\usepackage[skins,most]{tcolorbox}  
\usepackage{subcaption}
\usepackage{multicol}
\usepackage{stfloats}
\usepackage{algorithm}
\usepackage{amsmath}
\usepackage{algpseudocode}
\usepackage{amssymb}
\usepackage{mathtools}
\usepackage{mathtools}
\usepackage{graphics} 
\usepackage{epsfig} 
\usepackage{mathptmx} 
\usepackage{times} 
\usepackage{amsmath} 
\usepackage{amssymb}  
\usepackage{graphicx}
\usepackage{epstopdf}
\usepackage{pdfpages}
\usepackage{etoolbox}
\usepackage{amsbsy}
\usepackage{bm}
\usepackage{url}
\usepackage{accents}

\pdfoutput=1
\usepackage{hyperref}
\hypersetup{
    colorlinks=true,
    linkcolor=black,
    citecolor=black,
    filecolor=black,
    urlcolor=black,
}

\usepackage{tikz}
\usepackage[most]{tcolorbox}

\usepackage{cite}

\usepackage{stfloats}  
\def\horizontaldistance{\kern2pt}

\title{\LARGE \bf
Control of Legged Robots using Model Predictive Optimized Path Integral}
\author{Hossein Keshavarz$^{1}$, Alejandro Ramirez-Serrano$^{1}$, Majid Khadiv$^{2}$ 
\thanks{$^{1}$University of Calgary, Department of Mechanical Engineering, 2500 University Drive NW, Calgary, AB, Canada
Emails: {\tt\small {hossein.keshavarz},   {aramirez@ucalgary.ca}}}
\thanks{$^{2}$Munich Institute of Robotics and Machine Intelligence (MIRMI), Technical
University of Munich (TUM), Germany
Email: {\tt\small {majid.khadiv@tum.de}}}
}
\makeatletter
\newcommand*{\rom}[1]{\expandafter\@slowromancap\romannumeral #1@}
\makeatother
\begin{document}

\maketitle
\thispagestyle{empty}
\pagestyle{empty}
\begin{abstract}
Legged robots possess a unique ability to traverse rough terrains and navigate cluttered environments, making them well-suited for complex, real-world unstructured scenarios. However, such robots have not yet achieved the same level as seen in natural systems. Recently, sampling-based predictive controllers have demonstrated particularly promising results. This paper investigates a sampling-based model predictive strategy combining model predictive path integral (MPPI) with cross-entropy (CE) and covariance matrix adaptation (CMA) methods to generate real-time whole-body motions for legged robots across multiple scenarios. The results show that combining the benefits of MPPI, CE and CMA, namely using model predictive optimized path integral (MPOPI), demonstrates greater sample efficiency, enabling robots to attain superior locomotion results using fewer samples when compared to typical MPPI algorithms. Extensive simulation experiments in multiple scenarios on a quadruped robot show that MPOPI can be used as an anytime control strategy, increasing locomotion capabilities at each iteration.   
\end{abstract}
\section{INTRODUCTION}
\label{section:1}

Current algorithms for legged locomotion and manipulation are generally divided into two main groups: gradient-based model-predictive control (MPC)~\cite{wensing2023optimization} and gradient-free reinforcement learning (RL)~\cite{ha2024learning}. MPC policies do not require offline training as they leverage models for real-time trajectory optimization. However, MPC is computationally intensive at run-time, making it difficult to include whole-body dynamics and handle collision constraints in real-time. To reduce online computation, simplified models are often employed \cite{khadiv2020walking,daneshmand2021variable
,taouil2024non}, which prevents them from leveraging full-body dynamics to solve complicated locomotion or manipulation tasks. More recent works have enabled whole-body MPC and have shown promise in realizing complex real-world behaviors \cite{grandia2023perceptive,mastalli2023agile,meduri2023biconmp,li2024cafe,kim2025contact}. Furthermore, several works have reduced the online computation of whole-body MPC by learning the solutions \cite{carius2020mpc,viereck2022valuenetqp,khadiv2023learning,pua2024safe}.

Alternatively, sim-to-real RL has shown impressive progress in recent years \cite{bogdanovic2022model,miki2022learning,cheng2024extreme}, thanks to efficient parallel simulation on modern hardware~\cite{rudin2022learning,makoviychuk2021isaac} and improved sim-to-real transfer via domain randomization techniques~\cite{hwangbo2019learning,bogdanovic2022model,cheng2024extreme}. However, exploration in RL for legged robots remains a serious challenge, mainly due to the sparse rewards in complex environments \cite{zhang2024learning,dh2024diffusion}. Compared to MPC, one downside of RL policy optimization is the significant effort required to find suitable reward functions and a suitable set of locomotion parameters for each task. Furthermore, to successfully transfer learned control policies to real robots, an extensive sim-to-real randomization is required, and this procedure is entirely ad hoc and task dependent. 

Sampling-based MPC has recently emerged as an attractive middle-ground that does not require offline training and at the same time can leverage parallel simulation and mitigate issues of gradient-based methods in handling discontinuous and non-convex dynamics. 
Among different approaches, model predictive path integral (MPPI)~\cite{williams2017model}, covariance matrix adaptation (CMA)~\cite{hansen2016cma}, and cross-entropy method (CEM) \cite{de2005tutorial} have attracted a lot of attention, and several variants of them have been proposed.
Contrary to the common belief that sampling-based approaches are computationally intractable for high-dimensional systems, recent research has shown otherwise \cite{howell2022predictive,xue2024full,alvarez2024realtime,kurtz2025generative}. However, there is still no consensus among researchers on which technique is most suitable for complex legged locomotion control. This paper aims to investigate this.
In summary, this paper provides the following contributions:
\begin{enumerate}
    \item We present a comparative analysis of different sampling-based MPCs, underscoring the practical advantages and effectiveness of a combined strategy, namely model predictive optimized path integral (MPOPI), for locomotion and loco-manipulation problems. 
    \item We apply MPOPI to the problem of quadrupedal locomotion and loco-manipulation. To the best of our knowledge, this is the first work that demonstrates the application of such strategies in the context of locomotion.
    \item We carry out a wide range of simulation experiments demonstrating the capabilities of MPOPI for solving high-dimensional and complex tasks for generating whole-body motions in real time.
\end{enumerate}
This paper is structured as follows: Section~\ref{section:2} presents MPPI, CMA and CE methods. Section~\ref{section:3} describes the locomotion framework based on MPOPI, which is a combination of MPPI, CE and CMA, and its key components. In Sections~\ref{section:4}, the effectiveness of the locomotion framework based on MPOPI is demonstrated, highlighting its superiority over MPPI through an extensive comparison across various scenarios. In Section \ref{section:5}, the hyperparameters of MPOPI and their effects on the performance are further analyzed. Finally, Section~\ref{section:6} summarizes and concludes the paper.

\section{Preliminaries}
\label{section:2}
This section describes three widely adopted sampling-based MPC formulations from the literature: MPPI, CMA, and CE.
\subsection{Model predictive path integral (MPPI)}
MPPI samples $\it N$ possible control trajectories from a multivariate Gaussian distribution $u_t \sim \mathcal{N}(\mu_t, \Sigma_t)$, where $\mu_t$ is the mean and $\Sigma_t$ is the covariance matrix at time $\it{t}$. The cost of each sampled control trajectory is then evaluated by simulating these trajectories in parallel. The optimal control input to the system is then calculated through an exponentially weighted average of the samples based on their corresponding cost as follows \cite{williams2017model}:
\begin{subequations}
\begin{align}
    &\omega_n = \frac{\exp\left(-\frac{\mathcal{L}_n - \mathcal{L}_{\min}}{\lambda}\right)}{\sum_{n=1}^{N} \exp\left(-\frac{\mathcal{L}_n - \mathcal{L}_{\min}}{\lambda}\right)},\label{eq:1}\\ &\mu_t = \sum_{n=1}^{N} \omega_n u_t,
\label{eq:2}
\end{align}
\end{subequations}
where $\mathcal{L}_n$ represents the cost of the $\it n^{th}$ trajectory, $\omega_n$ denotes the corresponding trajectory weight, and $\lambda$ is the temperature parameter that determines the controller’s sensitivity to differences in trajectory cost. A lower value of $\lambda$ increases the influence of the best-performing trajectory, while a higher value distributes the weight more uniformly across all tested samples~\cite{williams2017model}. Such a process is repeated in a receding horizon fashion, and the mean is updated at each time step using a constant covariance matrix (i.e., $\Sigma_{t-1} = \Sigma_t$). The MPPI algorithm is summarized in Algorithm\hspace{0.05cm}\ref{alg:MPPI}.
\begin{algorithm}[]
\caption{MPPI Control Algorithm}
\label{alg:MPPI}
\textbf{Require:} Initial state $x_0$, control parameters $(\mu_0, \Sigma_0)$, number of samples $N$, time horizon $T$, temperature parameter $\lambda$ \\
\textbf{Ensure:} Optimal control input $u$
\begin{algorithmic}[1]
\State Initialize control trajectory mean $\mu_t$ and covariance $\Sigma_t$
\State Initialize sample trajectory cost $\mathcal{L}_n = 0$
\For{each sample $n = 1, \ldots, N$}
    \For{each timestep $t = 0, \ldots, T$}
        \State Sample action sequence: $u_t \sim \mathcal{N}(\mu_t, \Sigma_t)$
        \State Simulate forward for $T$ time steps using $u_n$
        \State Compute the resulting cost and add it to $\mathcal{L}_n$
    \EndFor
\EndFor
\State Compute weights $\omega_n$ for all samples using Equation\hspace{0.05cm}(\ref{eq:1}) 
\State Update control trajectory mean $\mu_t$ using Equation\hspace{0.05cm}(\ref{eq:2})
\State Select the control input for the system: $u = \mu_t[0]$
\State Apply control input $u$
\State Shift control mean: $\mu_t \leftarrow \text{shifted}(\mu_t)$
\end{algorithmic}
\end{algorithm}
\subsection{Covariance matrix adaptation (CMA)} 
In MPPI, the covariance matrix ($\Sigma_t$) is treated as a fixed parameter throughout the iterations. CMA~\cite{hansen2016cma} employs a multivariate Gaussian distribution as the search distribution and updates its parameters (mean and covariance) to generate superior solutions. Algorithm\hspace{0.05cm}\ref{alg:CMA} shows CMA control algorithm that computes the optimal control input by iteratively updating the control trajectory's mean and covariance using sampling and weighted feedback. Note that it is important to update the covariance matrix first as the covariance update must rely on the old $\mu_t$ (i.e. the one used to sample). Variants of the algorithm can be formulated depending on the weight computation strategy. The parameter $\alpha \in (0,1]$ acts as a learning rate that controls how much the new samples influence the updated parameters. A smaller $\alpha$ results in slower adaptation, relying more on the previous mean and covariance, which can help maintain stability but may slow convergence. Conversely, a larger $\alpha$ allows faster adaptation to recent sample information, which can speed up convergence but may introduce higher variance in the search process. 
\begin{algorithm}[]
\caption{CMA Control Algorithm}
\label{alg:CMA}
\textbf{Require:} Initial state $x_0$, control parameters $(\mu_0, \Sigma_0)$, number of samples $N$, time horizon $T$, learning rate $\alpha$ \\
\textbf{Ensure:} Optimal control input $u$
\begin{algorithmic}[1]
\State Initialize control trajectory mean $\mu_t$ and covariance $\Sigma_t$
\State Initialize sample trajectory cost $\mathcal{L}_n = 0$
\For{each sample $n = 1, \ldots, N$}
    \For{each timestep $\it{t} = 0, \ldots, T$}
        \State Sample action sequence: $u_t \sim \mathcal{N}(\mu_t, \Sigma_t)$
        \State Simulate forward for $T$ time steps using $u_n$
        \State Compute the resulting cost and add it to $\mathcal{L}_n$
    \EndFor
\EndFor
\State Compute weights $\it{w_{1}}, \ldots, \it{w_N}$ using Equation\hspace{0.05cm}(\ref{eq:3}) 
\State $\Sigma_t \leftarrow   \left(1 - \alpha \sum_{k=1}^{N} w_k\right)\Sigma 
  + \alpha \sum_{k=1}^{N} w_k (u_k - \mu_t)(u_k - \mu_t)^\top  $
  
\State $\mu_t \leftarrow \left(1 - \alpha \sum_{k=1}^{N} w_k\right)\mu_t 
  + \alpha \sum_{k=1}^{N} w_k u_k$
\State Select the control input for the system: $u = \mu_t[0]$
\State Apply control input $u$
\State Shift control mean: $\mu_t \leftarrow \text{shifted}(\mu_t)$
\end{algorithmic}
\end{algorithm}
In this paper, inspiring from~\cite{hansen2016cma}, the weights are computed using a logarithmic ranking scheme that assigns higher importance to top-ranked samples while gradually decreasing the weight for lower-ranked ones. Specifically, for each trajectory $i \in \{1, \dots, N\}$, the normalized weights are given by:
\begin{subequations}
\begin{align}
w_i &= \log\left(\frac{N+1}{i}\right), \quad i \in \{1, \dots, N\}, \label{eq:3}\\[4pt]
w_i^{\text{norm}} &= \frac{w_i}{\sum_{j=1}^N w_j + \epsilon},
\label{eq:4}
\end{align}
\end{subequations}
where $\epsilon$ is a small constant ($\texttt{1e-10}$) for numerical stability. This ranking-based weighting emphasizes better-performing samples while maintaining a smooth decay in importance across the ranked set.
\subsection{Cross entropy (CE)}
CE is a random sampling Monte Carlo method for importance sampling and optimization~\cite{pinneri2021sample}. Unlike MPPI and CMA, which consider all cost values when calculating sample weights, CE sorts the cost values and selects a fixed number of \emph{elite} sample candidates ($K_e$). This elite set is then used to determine the parameters of interest (i.e., mean and covariance) of the population for the next iteration. Consequently, unlike MPPI, where the covariance remains constant, the CE updates both the mean and the covariance at each time step as detailed in Algorithm\hspace{0.05cm}\ref{alg:CE}. In contrast to CMA, this method assigns a weight of $\frac{1}{K_e}$ to the best ${K_e}$ samples and zero to the others. Additionally, the learning rate is set to $\alpha = 1$. 
\begin{algorithm}[]
\caption{CE Control Algorithm}
\label{alg:CE}
\textbf{Require:} Initial state $x_0$, control parameters $(\mu_0, \Sigma_0)$, number of samples $N$, time horizon $T$ \\
\textbf{Ensure:} Optimal control input $u$
\begin{algorithmic}[1]
\State Initialize control trajectory mean $\mu_t$ and covariance $\Sigma_t$
\State Initialize sample trajectory cost $\mathcal{L}_n = 0$
\For{each sample $n = 1, \ldots, N$}
    \For{each timestep $\it{t} = 0, \ldots, T$}
        \State Sample action sequence: $u_t \sim \mathcal{N}(\mu_t, \Sigma_t)$
        \State Simulate forward for $T$ time steps using $u_n$
        \State Compute the resulting cost and add it to $\mathcal{L}_n$
    \EndFor
\EndFor
\State \textbf{Sort} ${\mathcal{L}}_{1:N}$ 
\State $\mu_t \leftarrow  \sum_{k=1}^{K_e} \frac{1}{K_e} u_k$

\State $\Sigma_t \leftarrow  \sum_{k=1}^{K_e} \frac{1}{K_e} (u_k - \mu_t)(u_k - \mu_t)^\top  $
\State Select the control input for the system: $u = \mu_t[0]$
\State Apply control input $u$
\State Shift control mean: $\mu_t \leftarrow \text{shifted}(\mu_t)$
\end{algorithmic}
\end{algorithm}

\section{Proposed locomotion framework}
\label{section:3}
In this section, the locomotion framework developed based on MPOPI is presented. Firstly, the MPOPI controller is described, followed by an explanation of how it is integrated into the developed locomotion framework.
\subsection{Model predictive optimized path integral (MPOPI)}
MPOPI combines the MPPI with CE and CMA methods to take advantage of each method with the goal of creating a unified approach that reduces the time needed to find a control action~\cite{asmar2023model}.
\begin{figure*}[tb]
\centering
\includegraphics[scale=0.75, trim={0.0cm 10.0cm 0.5cm 9.0cm}, clip]{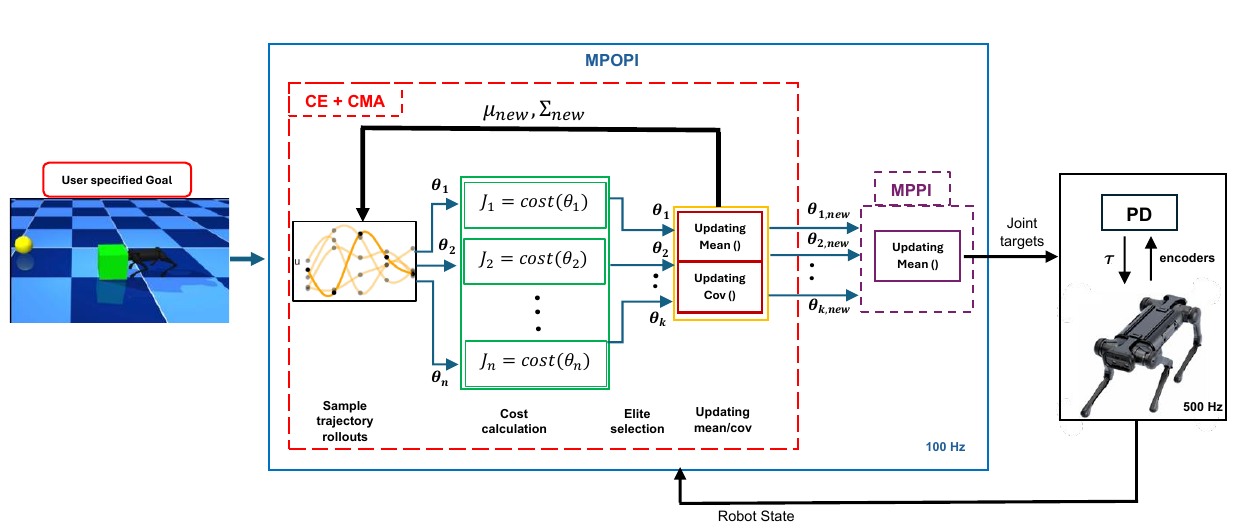}
\caption{Block diagram of the MPOPI locomotion control method.}
\label{fig:New Flowchart}
\end{figure*}
In contrast to MPPI, CMA and CE, in MPOPI an initial set of randomized samples is generated using a random or predefined mean and covariance values. Subsequently, such two parameters are updated via CE and CMA from where a new set of improved randomized samples is generated using the updated mean and covariance as detailed in Algorithm \ref{alg:MPOPI}.
The updating process is repeated for $L$ cycles, from which a new subset of improved robot motion trajectories (joint positions) is generated. Such a strategy enables an efficient search over actions to quickly find the optimal sequence, with reduced computation. When $L = 1$, the MPOPI algorithm provides similar results to the MPPI algorithm because the mean and covariance are not updated. 
\vspace*{5pt}
\begin{algorithm}[]
\caption{MPOPI Control Algorithm}
\label{alg:MPOPI}
\begin{algorithmic}[1]
\Require Initial state $x_0$, control parameters $(\mu_0, \Sigma_0)$, number of samples $N$, time horizon $T$, temperature parameter $\lambda$, updating cycle $L$, learning rate $\alpha$
\Ensure Optimal control input $u$
\State Initialize control trajectory mean $\mu_t$ and covariance $\Sigma_t$
\State $\mu_t' \gets \mu_t$
\State $\Sigma_t' \gets \Sigma_t$
\State Initialize sample trajectory cost $\mathcal{L}_n = 0$
\For{each $l = 1, \dots, L$}
  \For{each sample $n = 1, \dots, N/L$}
    \State Sample action sequence: $u_t \sim \mathcal{N}(0, \Sigma_t')$
    \For{each timestep $t = 0, 1, \dots$}
      \State Simulate forward for $T$ time steps using $u_n$
      \State Compute the resulting cost and add it to $\mathcal{L}_n$
      \State $\mathcal{L}_n += \lambda (1 - \alpha) \mu_t' \Sigma_t^{-1} (u_t + \mu_t' - \mu_t)$
    \EndFor
  \EndFor
  \If{$l < L$} 
      \State // \textcolor{red}{\textbf{CMA + CE Methods}}
      \State \textbf{Sort} ${\mathcal{L}}_{1:N/L}$ and select ${K_e}$ samples
      \State Compute weights $\it{w_{1}}, \ldots, \it{w_{K_e}}$ using Equation\hspace{0.05cm}(\ref{eq:3}) 
\State $\begin{aligned}
\Sigma_t' &\leftarrow \left(1 - \alpha \sum_{k=1}^{{K_e}} w_k\right)\Sigma_t' \\
&\quad + \alpha \sum_{k=1}^{{K_e}} w_k (u_k - \mu_t)(u_k - \mu_t)^\top
\end{aligned}$

\State $\mu_t' \leftarrow \left(1 - \alpha \sum_{k=1}^{{K_e}} w_k\right)\mu_t' 
+ \alpha \sum_{k=1}^{{K_e}} w_k u_k$
  \EndIf
\EndFor
\State // \textcolor{teal}{\textbf{MPPI Method}}
\State Compute weights $\omega_n$ for all samples using Equation\hspace{0.05cm}(\ref{eq:1}) 
\State Update control trajectory mean $\mu_t = \omega_n (u_t + \mu_t' - \mu_t)$
\State Select the control input for the system: $u = \mu_t[0]$
\State Apply control input $u$
\State Shift control mean: $\mu_t \gets \text{shifted}(\mu_t)$
\end{algorithmic}
\end{algorithm}

\subsection{Locomotion control using MPOPI}
In this work, we use the locomotion framework developed in~\cite{alvarez2024realtime} as a base framework. Our framework incorporates additional elements for updating the mean and covariance of the samples. A block diagram of the proposed system is illustrated in Fig. \ref{fig:New Flowchart}.  
The process starts with a user-specified goal, which defines the target for the robot. Within the MPOPI module, the framework samples a set of trajectory rollouts in parallel using MuJoCo physics engine~\cite{todorov2012mujoco} on a multi-core CPU, each parameterized by control sequences $\theta_{1}, \theta_{2}, ..., \theta_{n}$. These trajectories are evaluated using a cost function, and the best-performing (elite) trajectories ($\theta_{1}, \theta_{2}, ..., \theta_{k_e}$) are selected. The mean and the covariance of the distribution are updated based on the elite set using the CE method. This process repeats $L$ cycles and the joint targets input to the system are then calculated through
an exponentially weighted average of the samples using MPPI. The resulting joint targets are passed to a Proportional-Derivative (PD) controller, which operates at 500 Hz to compute torque commands and send them to the robot. 
Following~\cite{howell2022predictive}, this approach offers two key advantages: $\it{i)}$ it reduces the size of the search space, and $\it{ii)}$ promotes the generation of smoother control inputs. 

\section{Simulation Results}
\label{section:4}

In this section, we use the simulation of a Go1 quadruped robot, controlled by MPOPI and compare its performance against MPPI. For a fair comparison, while MPPI uses 30 samples, MPOPI uses 10 samples repeated over 3 cycles (L = 3), resulting in the same total number of samples. The other optimization variables, such as the time horizon (T = 40) and the temperature parameter ($\lambda = 0.1$), are kept the same for both methods. To compare, the robot was commanded to perform the following tasks:
\begin{itemize}
\item Climbing stairs,
\item Climbing over boxes of different heights, and
\item Pushing a box to a user-defined position. 
\end{itemize}

\subsection{Stair Climbing}
\begin{figure}[tb]
\centering
\includegraphics[scale=0.5, trim ={3.0cm 14.0cm 3.0cm 3.9cm},clip]{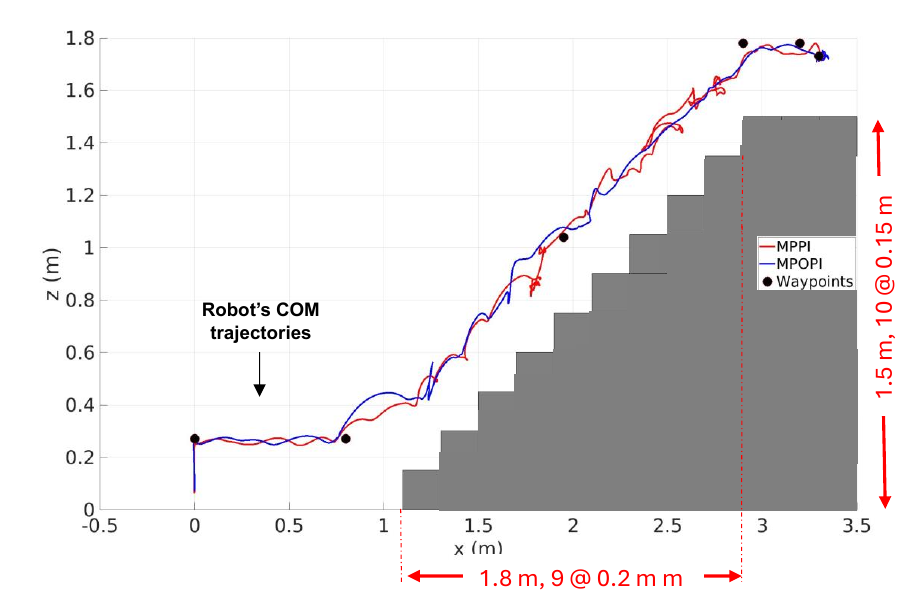}
\caption{Stair climbing (side view).}
      \label{fig:Stair_climbing_side}
\end{figure}

Stair climbing requires great joint exploration. To test the abilities of MPOPI to deal with such climbing scenarios, diverse tests were performed asking the robot to climb stairs of different heights (e.g., 15 cm) and tread length (e.g., 20 cm) as illustrated in Fig. \ref{fig:Stair_climbing_side}. 

\begin{figure}[tb]
\centering
\includegraphics[scale=0.13, trim ={3.0cm 1.0cm 3.5cm 15.0cm},clip]{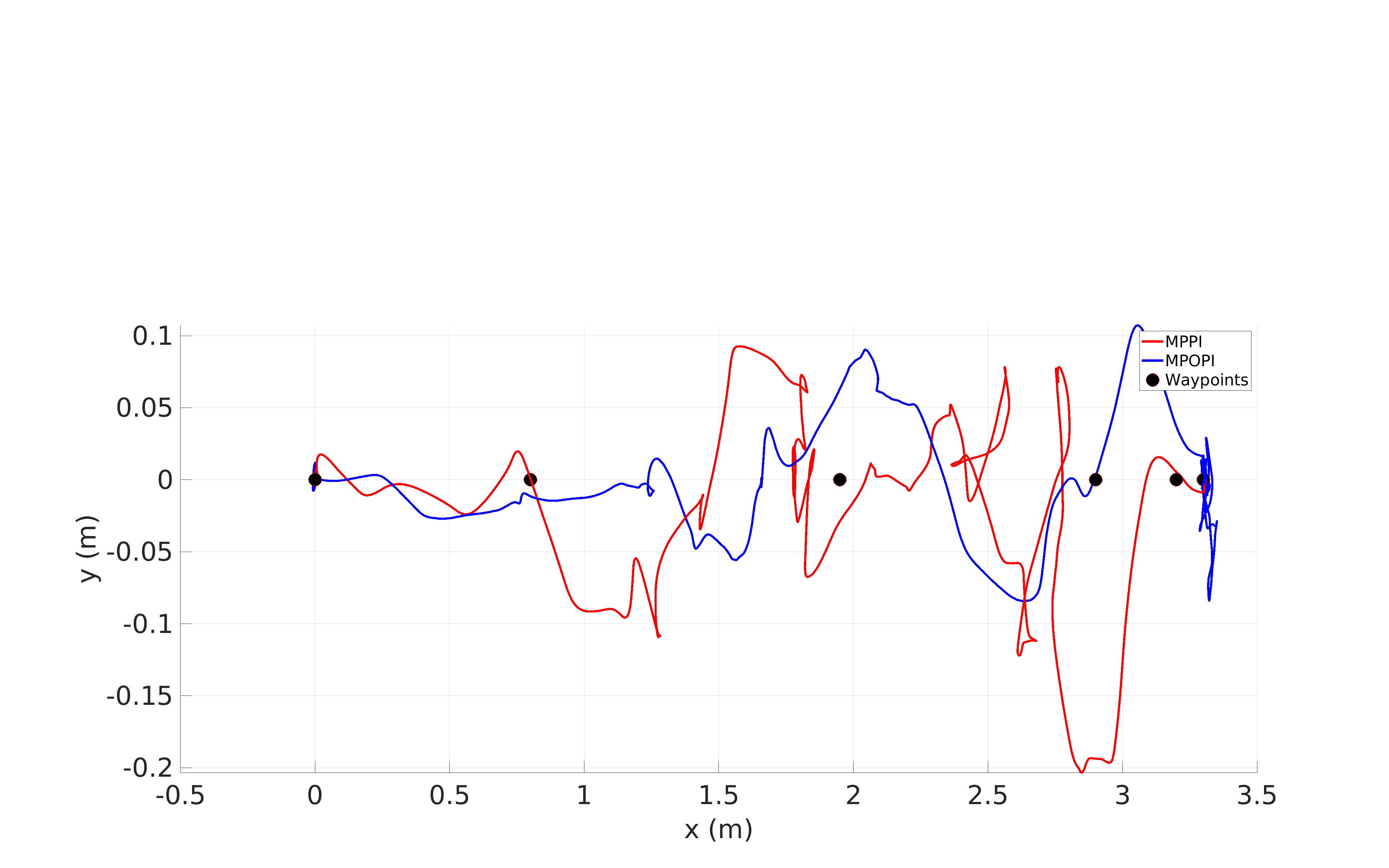}
\caption{Stair climbing (top view).}
      \label{fig:Stair_climbing_top}
\end{figure}

In this scenario, the robot was commanded to achieve a set of \emph{m} waypoints located along the stairs while asked to achieve a given (user-defined) locomotion speed. The obtained results (Fig. \ref{fig:Stair_climbing_top}) show that MPOPI achieves the most stable and accurate performance as defined by the metrics of center of mass (CoM) trajectory smoothness and position error. The robot achieves a smooth trajectory with minimal lateral deviation, while MPPI shows some greater trajectory fluctuations (Fig. \ref{fig:Stair_climbing_top}). MPOPI maintains a smooth and consistent ascent while tracking elevation changes effectively. 

\begin{figure*}[!t]
\centering
\includegraphics[scale=0.85, trim={4.5cm 14.5cm 1.0cm 3.0cm}, clip]{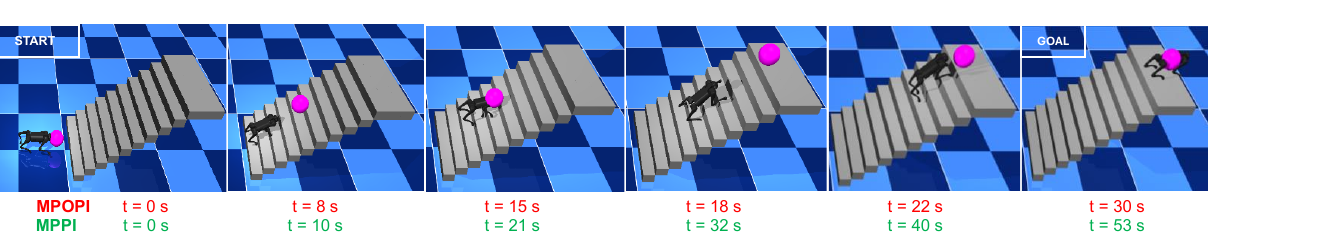}
\caption{Stair climbing using MPOPI and MPPI sampling-based MPCs.}
\label{fig:stair}
\end{figure*}

Compared to MPPI, the MPOPI approach is able to generate effective robot motion strategies in less time. As an example, Fig. \ref{fig:stair} compares the time required for a quadruped robot to find a suitable whole-body locomotion using MPOPI (in red) and MPPI (in green). MPOPI demonstrates a significantly faster ascent, where the robot takes about half the time (30 s) to complete the same task when compared to the time taken when using MPPI (i.e., 53 s).

\begin{figure}[tb]
\centering
\includegraphics[scale=0.13, trim ={3.0cm 0.0cm 3.0cm 2.0cm},clip]{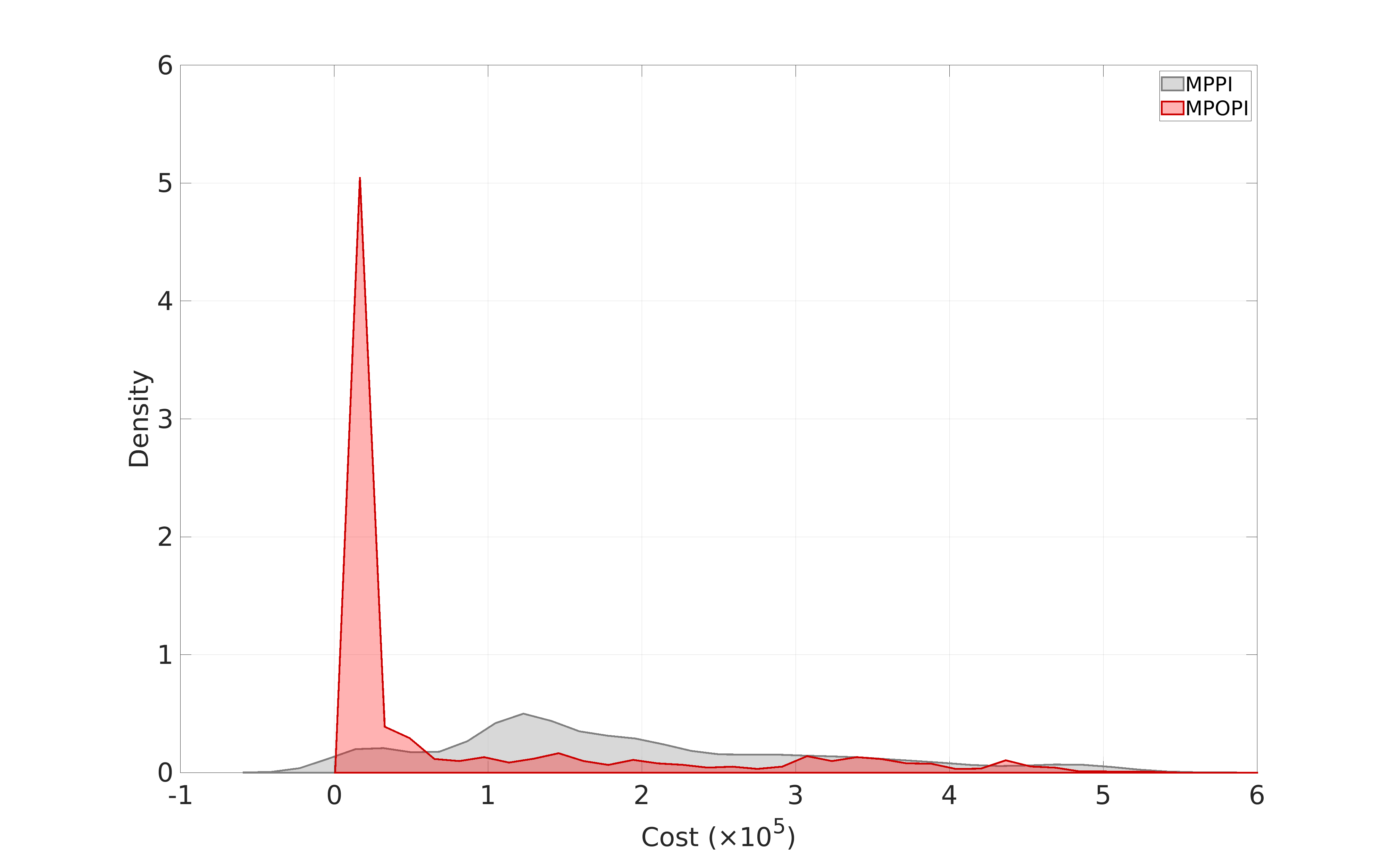}
\caption{The cost distribution of MPOPI and MPPI in the stair climbing scenario. In this setting, we can observe that MPPI has flattened distributions, while MPOPI's cost distribution has a sharp peak.}
      \label{fig:cost_density_stairs}
\end{figure}
The obtained cost density for this specific example (Fig. \ref{fig:cost_density_stairs}) indicates a superior performance of MPOPI. Its sharp and narrow peak near the lower end of the cost axis indicates both low average cost and low variability across trials. MPPI shows higher and more inconsistent costs, with a widely spread and relatively flat distribution. 

\subsection{Climbing large boxes}
Similar to stair climbing, MPOPI was analyzed on enabling robots to climb over large boxes (i.e., obstacles having larger risers and treads when compared to those found in stairs). Two different box heights were tested in simulations: 40 cm and 45 cm. As the box height increases, the climbing task becomes more difficult, requiring the robot to perform greater exploration in the joint space. 

\begin{figure}
\centering
\begin{subfigure}[b]{0.9\linewidth}
    \centering
    \includegraphics[width=\linewidth, trim={3.0cm 0.0cm 3.0cm 2.0cm}, clip]{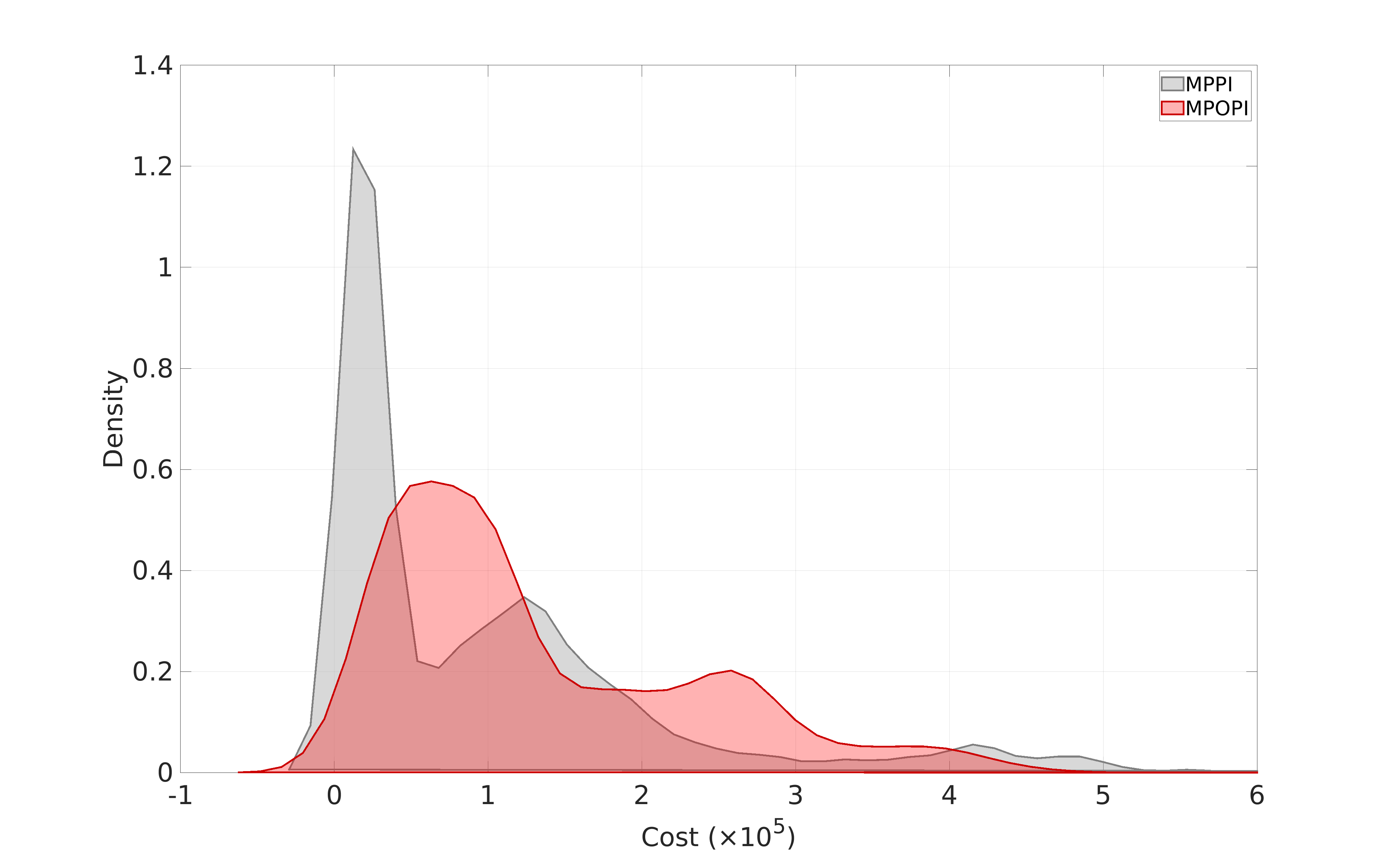}
    \caption{Climbing a box when its height is 40 cm.}
    \label{fig:cost_density_40}
\end{subfigure}
\vspace{0.5em} 
\begin{subfigure}[b]{0.9\linewidth}
    \centering
    \includegraphics[width=\linewidth, trim={3.0cm 0.0cm 3.0cm 2.0cm}, clip]{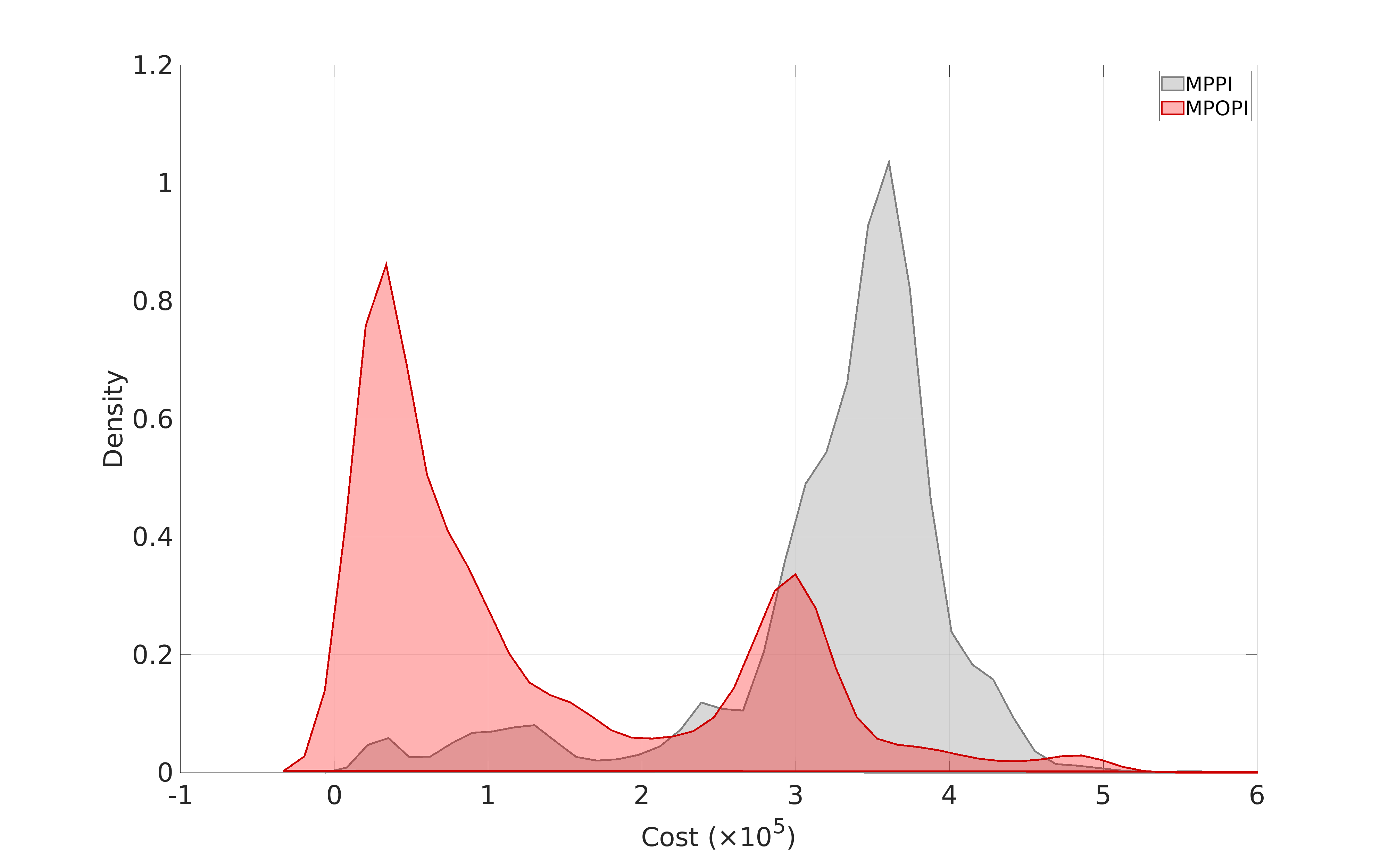}
    \caption{Climbing a box when its height is 45 cm.}
    \label{fig:cost_density_45}
\end{subfigure}
\caption{The cost distribution of MPOPI and MPPI in the third scenario: climbing a box. MPPI's distribution flattens with the taller box. In contrast, MPOPI maintains a sharp peak in both situations.}
\label{fig:cost_density_all}
\end{figure}

Figure \ref{fig:cost_density_40} shows that when the robot (having a nominal body height of 27 cm when walking) climbs obstacles with a height of 40 cm, the task is done successfully using both methods. Both MPPI and MPOPI have a low cost, while MPOPI shows a high density near the lower cost range with more spread, reflecting greater variability in cost. When increasing the obstacle's height (e.g., 45 cm), MPOPI shows a prominent, sharp peak near the lower cost range, indicating its ability to perform efficiently and consistently, even with increased task difficulty as shown in Fig. \ref{fig:cost_density_45}. 
Meanwhile, MPPI exhibits a broader and flatter distribution, with more mass spread across higher cost values, suggesting reduced efficiency and increased variability in performance. Ultimately, MPPI was unable to complete the task successfully.


\subsection{Pushing a box}
In this scenario, the controllers are evaluated based on their whole-body contact planning capabilities for pushing a box from a predefined initial position to a final position (i.e., locomanipulation tasks). Note that the tests focus on the position of the box along the trajectory (and not on its orientation). During the tests, the box size was constrained to a size of 0.36m × 0.36m × 0.36m with a mass of 3.5 kg. The box is initially positioned at 1m in front of the quadruped, and the task is to bring the position of the box within 0.2 m of the goal waypoints at any time. The robot is requested to push the targeted box with its frontal face following a sequence of two 2D waypoint positions, i.e., the robot is allowed to push the box at any point on the box's faces. 
When the initial position of the robot and the two desired waypoints of the box define a linear trajectory (i.e., pushing the box forward), both MPPI and MPOPI controllers achieve a smooth linear trajectory and achieve the goal. Both controllers show small trajectory errors.
\begin{figure}[tb]
\centering
\includegraphics[scale=0.8, trim ={3.0cm 12.0cm 1.0cm 8.0cm},clip]{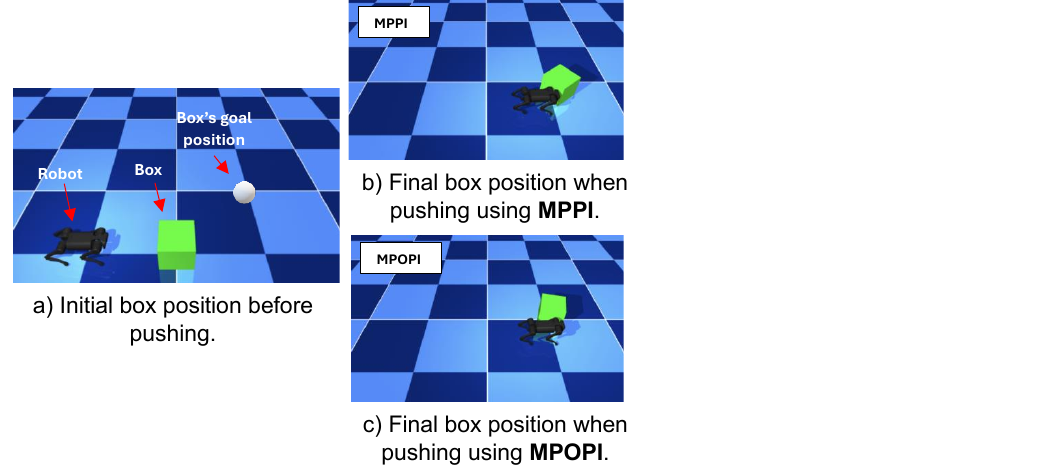}
\caption{Comparison of MPPI and MPOPI methods for pushing a box.}
      \label{fig:push_comp}
\end{figure}

\begin{figure}[tb]
\centering
\includegraphics[scale=0.8, trim ={5.0cm 9.0cm 3.0cm 9.0cm},clip]{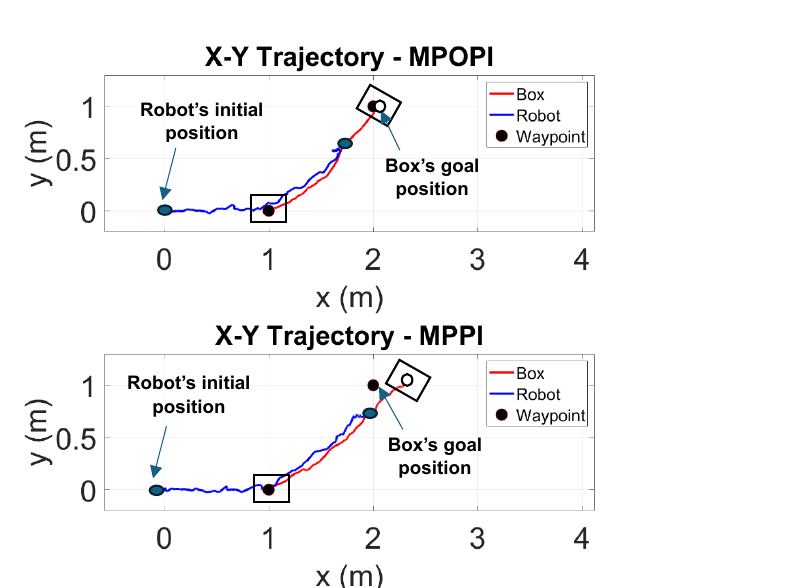}
\caption{Trajectories of the robot and the box in the XY-plane.}
      \label{fig:push_lateral}
\end{figure}

However, pushing the box along a 3-point trajectory that does not define a rectilinear path presents more challenges, resulting in different performance between the MPPI and the MPOPI controllers as shown in Fig. \ref{fig:push_comp}. Also, Fig. \ref{fig:push_lateral} demonstrates that MPOPI enables smoother and more direct trajectories for both the robot and the box, resulting in the box reaching its goal with greater accuracy. In contrast, MPPI exhibits a slightly irregular trajectory, particularly in the final stage, where the box deviates from reaching the final goal destination. Additionally, the robot's path under MPPI shows greater lateral oscillation, suggesting less efficient coordination between the robot's motion and the box's trajectory. These results indicate that updating the mean and the covariance of samples enables better adaptation of the sampling distribution, leading to improved motion control performance in complex, directionally coupled tasks. 

\begin{figure}[tb]
\centering
\includegraphics[scale=0.75, trim ={4.0cm 8.0cm 5.0cm 10.0cm},clip]{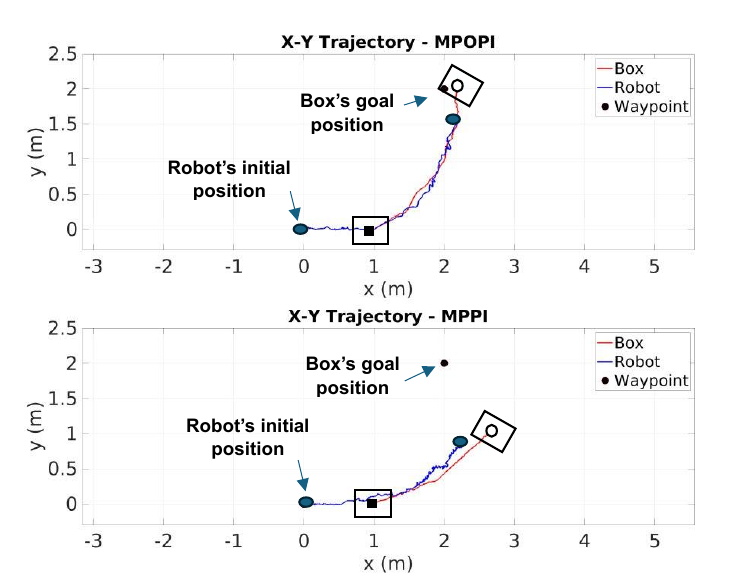}
\caption{Trajectories of the robot and the box toward a far target position in the XY-plane.}
      \label{fig:push_far}
\end{figure}

In another test, the goal for the box was set at (2, 2), which is farther away compared to the previous test. Figure \ref{fig:push_far} illustrates the X–Y trajectories of the robot and the box for both controllers. In the MPOPI case (top), the robot successfully pushed the box to the target location, closely following a direct path with minimal deviation. In contrast, MPPI (bottom) was unable to reach the target, with both the robot and the box deviating significantly from the desired trajectory. This demonstrates MPOPI’s superior ability to maintain task accuracy over longer distances compared to MPPI.


\section{MPOPI key Parameters}
\label{section:5}
Based on the simulation results, the MPOPI controller delivers superior performance in locomotion and loco-manipulation tasks. Overall, the method performs the best in complex environments where robots require greater exploration in the joint space to identify a viable path. However, two key parameters in the MPOPI controller (Algorithm \ref{alg:MPOPI}) have been observed to significantly influence its performance: $\it{i)}$ the number of cycles performed to update the mean and the covariance (denoted as $L$), and $\it{ii)}$ the covariance matrix updating process during optimization.

\begin{figure}
\centering
\includegraphics[scale=0.14, trim ={3.0cm 1.0cm 3.0cm 1.0cm},clip]{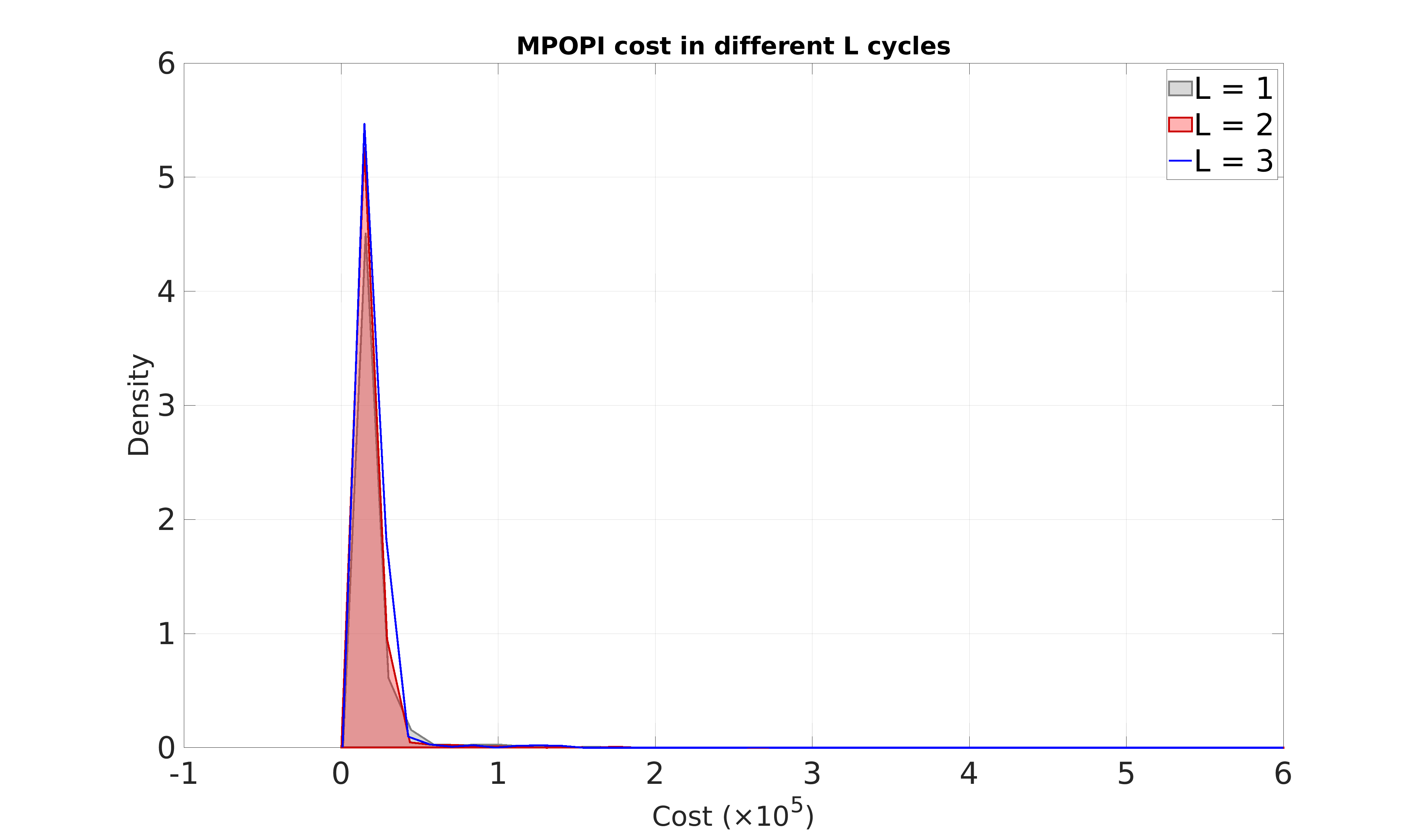}
\caption{Policy rollout costs for walking while changing $L$.}
      \label{fig:changing_L_walking}
\end{figure}

\subsection{Number of loops ($L$)}
\begin{figure}
\centering
\includegraphics[scale=0.14, trim ={3.0cm 1.0cm 3.0cm 1.0cm},clip]{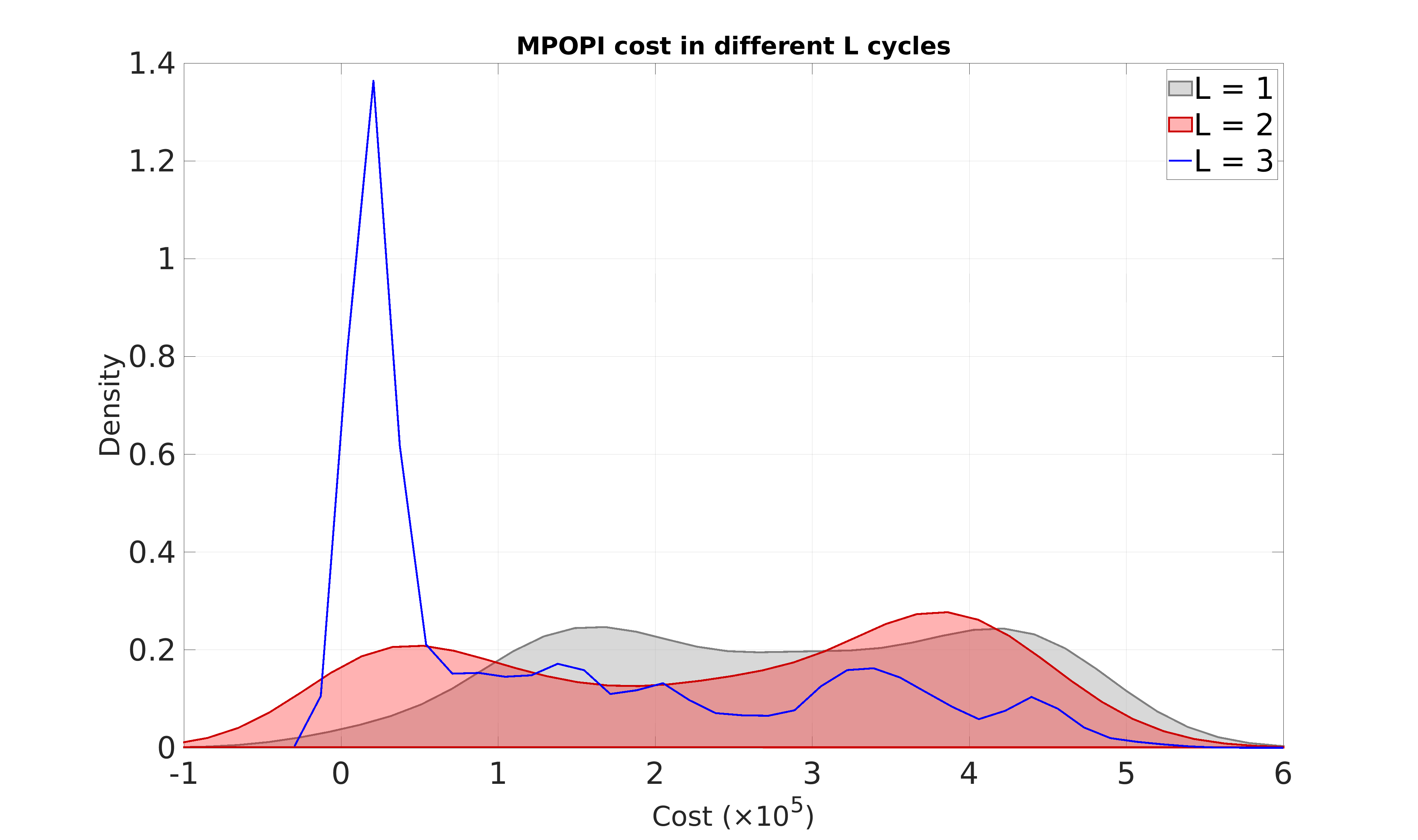}
\caption{Policy rollout costs for stair climbing while changing $L$.}
      \label{fig:changing_L_stair}
\end{figure}

To determine the best set of parameters, diverse methodologies can be used, including running diverse tests from which the mean and standard deviations of each test can be computed. Figs \ref{fig:changing_L_walking} and \ref{fig:changing_L_stair} show examples for two illustrative scenarios for a total of 30 samples (N = 30): walking and stair climbing. Fig. \ref{fig:changing_L_walking} shows that increasing $L$ does not affect the cost distribution, indicating that adjusting $L$ for walking offers no benefit.  However, for more complex tasks (typically requiring more exploration) such as stair climbing, having an $L$ larger than one ($L>1$) is preferred (Fig. \ref{fig:changing_L_stair}). 

\begin{figure}[tb]
\centering
\includegraphics[scale=0.14, trim ={3.0cm 1.0cm 3.0cm 1.0cm},clip]{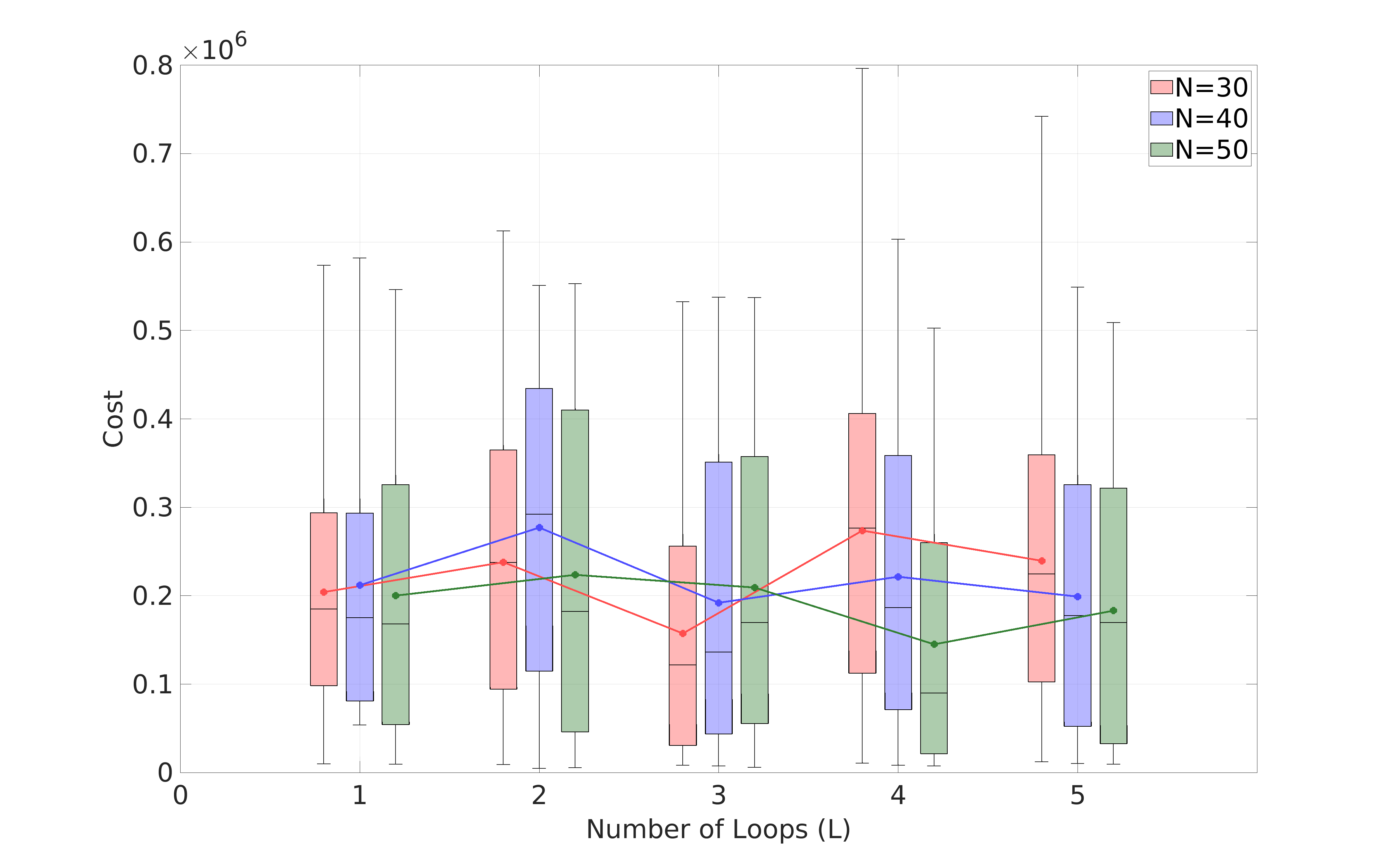}
\caption{Policy rollout costs for stair climbing while changing $L$.}
      \label{fig:changing_L_stair1}
\end{figure}

Figure \ref{fig:changing_L_stair1} illustrates the effect of the number of loops $L$ across different sample sizes (N = 30, 40, 50). These observations indicate that in different number of samples, larger than one cycle is preferred. 
\subsection{Covariance matrix}
While MPPI uses a fixed initial covariance matrix $\Sigma = \lambda_{\text{ini}} I_{12}$, where $\lambda_{\text{ini}}$ represents the standard deviation for each joint, MPOPI updates the covariance matrix $\Sigma$ at each time step. This involves computing the eigenvalues and eigenvectors of $\Sigma$, applying a lower bound to the eigenvalues to ensure numerical stability, and reconstructing the updated covariance matrix using the bounded eigenvalues and their corresponding eigenvectors. To analyze the effect of updating $\Sigma$, an \emph{exploration magnitude} is defined as the largest eigenvalue $\lambda$ of the covariance matrix. 

\begin{figure}
\centering
\includegraphics[scale=0.55, trim ={2.5cm 10.0cm 3.0cm 9.0cm},clip]{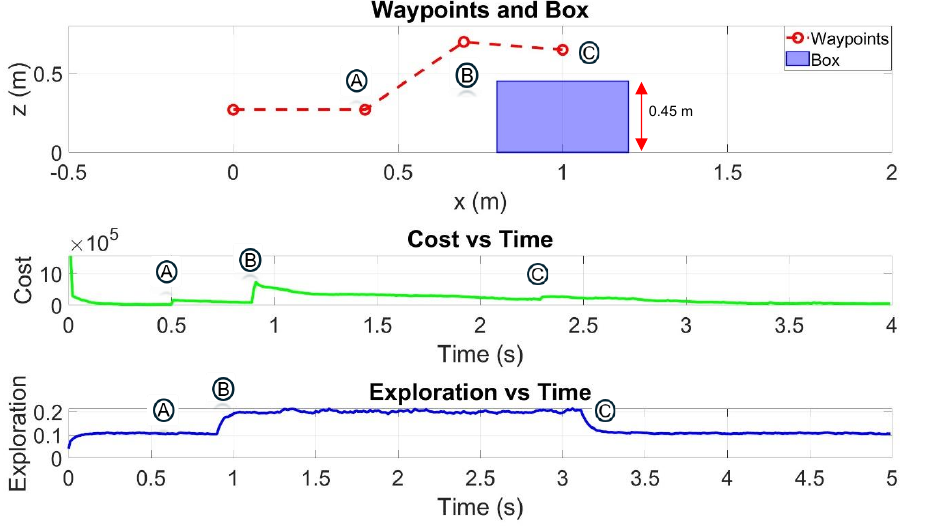}
\caption{Exploration magnitude and cost distribution for climbing a big box.}
      \label{fig:exploration}
\end{figure}

Figure \ref{fig:exploration} shows how the exploration magnitude and the cost value change when the robot climbs a tall box. This task is very hard to achieve when MPPI  is used, and the robot fails. However, MPOPI enables the robot to complete the task.  
As shown in Fig. \ref{fig:exploration}, the task can be divided into three steps. In the first step, the robot walks from the initial position to \textcircled{\footnotesize A} where the exploration magnitude is at a minimum. For the second step (\textcircled{\footnotesize A} - \textcircled{\footnotesize B}), the robot starts climbing the box, stimulating an increased value of the exploration magnitude. Here, as the cost value decreases, the exploration magnitude remains large, enabling the robot to adapt to the challenging task. At \textcircled{\footnotesize C}, the exploration magnitude stabilizes and decreases, suggesting the robot has found an effective policy and no longer requires significant variability in action sampling. This coordinated behavior across plots demonstrates the effectiveness of adaptive exploration in the MPOPI framework for complex motion planning tasks.

\section{Conclusions}
\label{section:6}
In this paper, we developed a locomotion controller based on MPOPI and applied it to several quadrupedal locomotion and loco-manipulation tasks. We conducted an extensive set of simulation experiments to compare the performance of MPOPI to other widely used sampling-based MPC techniques, i.e., MPPI. Our results show that MPOPI provides better performance across a number of performance indices, including tracking error, and provides lower costs when compared to the traditional MPPI for generating real-time whole-body motion plans. We showed that while MPOPI shows similar performance to MPPI when the number of loop iterations $L$ of MPOPI is one, it outperforms MPPI when $L>1$, resulting in a more versatile formulation for whole-body MPC for legged robots that can be applied to robots operating in complex unstructured environments or performing complex tasks where MPPI tends to fail.  
In the future, we plan to apply our developed locomotion control framework based on MPOPI on real hardware.
\bibliography{bibliography}
\bibliographystyle{IEEEtran}

\end{document}